%% file: main.tex
\documentclass[10pt,twocolumn,letterpaper]{article}

\usepackage{cvpr} 
\usepackage[ruled,vlined,linesnumbered]{algorithm2e}
\usepackage{amsmath,amssymb}
\usepackage{cuted}
\usepackage{algpseudocode}  
\usepackage{amsmath}        
\usepackage{amssymb}        
\usepackage{multirow}
\usepackage{booktabs}
\usepackage[table]{xcolor}
\input{preamble}
\definecolor{cvprblue}{rgb}{0.21,0.49,0.74}
\usepackage[pagebackref,breaklinks,colorlinks,allcolors=cvprblue]{hyperref}
\usepackage{float}
\usepackage{dsfont}
\usepackage[ruled,vlined]{algorithm2e}
\def\paperID{9658} 
\def\confName{CVPR}
\def\confYear{2026}

\title{AR$^2$-4FV: Anchored Referring and Re-identification for Long-Term Grounding in Fixed-View Videos}

\author{Teng Yan\thanks{Equal contribution.}, Yihan Liu\footnotemark[1], Jiongxu Chen, Teng Wang, Jiaqi Li, Bingzhuo Zhong\thanks{Corresponding author.}\\
The Hong Kong University of Science and Technology (Guangzhou)\\
{\tt\small \{tyan497,yliu135,jchen074, twang560,jli657,bingzhuoz\}@connect.hkust-gz.edu.cn}
}

\begin{document}
\maketitle
\begin{strip}
\vspace{-1.3cm}
  \centering
\includegraphics[width=\textwidth]{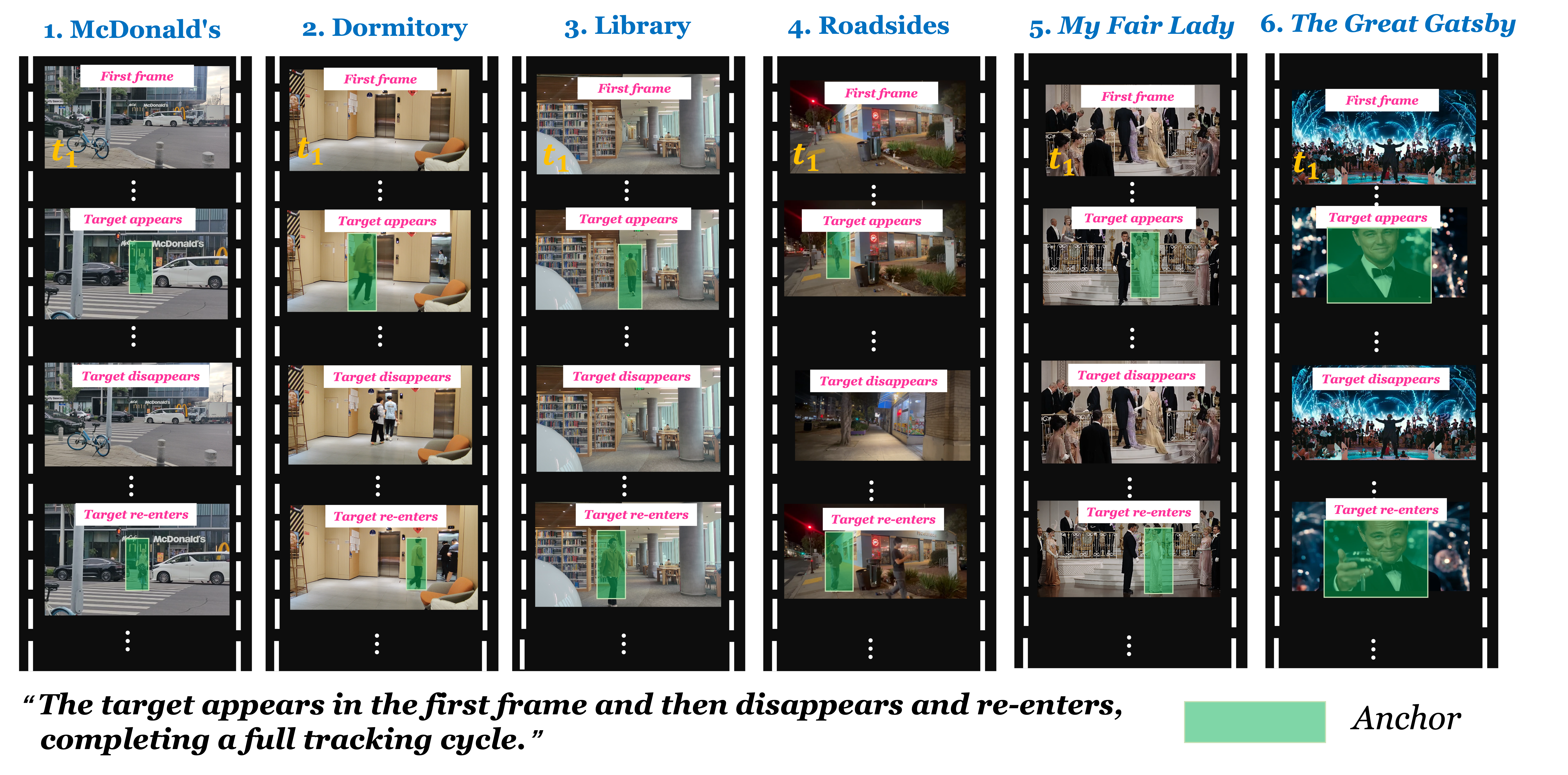}
  \captionof{figure}{\textbf{AR$^2$-4FV-Bench scenes and protocol.} Panels from several fixed-view locations and two cinematic sequences. For each sequence we show the first frame, a visible instance, a period of absence, and the re-entry moment, constituting a full tracking cycle. Green overlays indicate anchors (stable background supports). The benchmark targets long-term disappearance and re-entry under a fixed camera and evaluates identity-consistent trajectories without assuming the target is visible at first frame.}
  \label{fig:fig1}
\end{strip}

\input{sec/0_abstract}    
\input{sec/1_intro}

\input{sec/2_relatedwork}
\input{sec/3_Dataset}

\input{sec/4_Methodology}
\input{sec/5_Experiments}

\input{sec/6_Conclusion}

\input{sec/7_Acknowledgement}
{
    \small
    \bibliographystyle{ieeenat_fullname}
    \bibliography{main}
}


\end{document}

%% file: sec/0_abstract.tex
\begin{abstract}
Long-term language-guided referring in fixed-view videos is challenging: the referent may be occluded or leave the scene for long intervals and later re-enter, while framewise referring pipelines drift as re-identification (ReID) becomes unreliable. AR$^2$-4FV leverages background stability for long-term referring. An offline Anchor Bank is distilled from static background structures; at inference, the text query is aligned with this bank to produce an Anchor Map that serves as persistent semantic memory when the referent is absent. An anchor-based re-entry prior accelerates re-capture upon return, and a lightweight ReID-Gating mechanism maintains identity continuity using displacement cues in the anchor frame. The system predicts per-frame bounding boxes without assuming the target is visible in the first frame or explicitly modeling appearance variations. AR$^2$-4FV achieves +10.3\% Re-Capture Rate (RCR) improvement and –24.2\% Re-Capture Latency (RCL) reduction over the best baseline, and ablation studies further confirm the benefits of the Anchor Map, re-entry prior, and ReID-Gating.
\end{abstract}

%% file: sec/1_intro.tex
\section{Introduction}
\label{sec:intro}

Language-guided referring in videos enables users to refer targets via text query without needing pre-registered IDs\cite{Bridge,ReferFormer,CLUE,SgMg,SSA}. It is especially valuable for real-world applications such as public surveillance, intrusion detection, and long-term behavior analysis where fixed-view cameras are widely deployed. Fixed-view videos provide consistent spatial layouts and persistent environmental references, but robust long-term referring in such settings remains difficult due to occlusion, disappearance, re-entry, and appearance degradation. Unlike short-term referring tasks\cite{Fully,Losh,SOC}, long-term scenarios suffer from semantic memory loss when the target is absent for extended periods, making re-capture upon re-entry challenging. Over long durations, appearance features also become unreliable under illumination changes, pose variations, and environmental factors\cite{li2019}, leading to identity drift. Although the stable background structure in fixed-view scenes offers a strong spatial prior, effectively integrating it with appearance features to preserve identity consistency across long-term videos is still technically challenging. 

In long-term fixed-view scenarios, framewise or short-window association approaches tend to drift, as their semantic memory is easily disrupted by near-semantic distractors\cite{videnovic,survey}. When the target is occluded or leaves the field of view for an extended period, these methods struggle to preserve the text–referent consistency upon re-entry. ReID-based strategies\cite{Seas,Instruct,Deepchange} rely primarily on appearance features making identity association prone to drift. Moreover, existing language-guided models rarely incorporate spatial priors to mitigate the effects of invisibility, making them without a query-conditioned reference that persists when the referent is absent. Consequently, maintaining the text–referent alignment throughout long-term disappearance and re-entry remains a key but unresolved challenge.

To address these issues, AR$^2$-4FV is proposed as a framework that couples the referring expression with the invariant background structures in fixed-view videos. Offline, Anchor Bank is distilled from static background regions. Online, the text query is aligned with this bank to generate an Anchor Map.
This map maintains text–referent correspondence via language-to-scene alignment and provides an anchor-based re-entry prior to guide efficient and stable re-capture once the target re-enters. In addition, a lightweight ReID-Gating module maintains identity continuity by jointly validating candidate appearance, anchor evidence and displacement in the anchor coordinate space. Unlike generic long-term tracking, we do not aim to handle cross-camera outfit changes; instead, we focus on exploiting intra-scene stability in fixed-view videos. 
AR$^2$-4FV enables robust long-term referring in fixed-view videos without assuming the referent is visible in the first frame. To support systematic evaluation under this setting, we additionally introduce AR$^2$-4FV-Bench, a dedicated benchmark for long-term referring with explicit disappearance, occlusion and re-entry cases.

Our main \textbf{contributions} are summarized as follows: 
\begin{itemize}




    \item \textbf{AR$^2$-4FV.} A framework for long-term language-guided referring and re-identification in fixed-view videos without assuming initial visibility.
	\item \textbf{Language-anchored scene memory.} Introducing an offline Anchor Bank and an online Anchor Map to form a query-conditioned spatial prior, combined with an anchor-based re-entry prior and ReID-Gating to maintain identity continuity.
	\item \textbf{AR²-4FV-Bench.} A fixed-view, long-term referring and re-identification benchmark with language-to-scene alignment, framewise visibility and re-entry annotations.
\end{itemize}

%% file: sec/2_relatedwork.tex
\section{Related Work}
\label{sec:related work}
\begin{figure*}[t]
    \centering
    
    \begin{subfigure}[t]{0.3\textwidth}
        \centering
        \includegraphics[width=\linewidth]{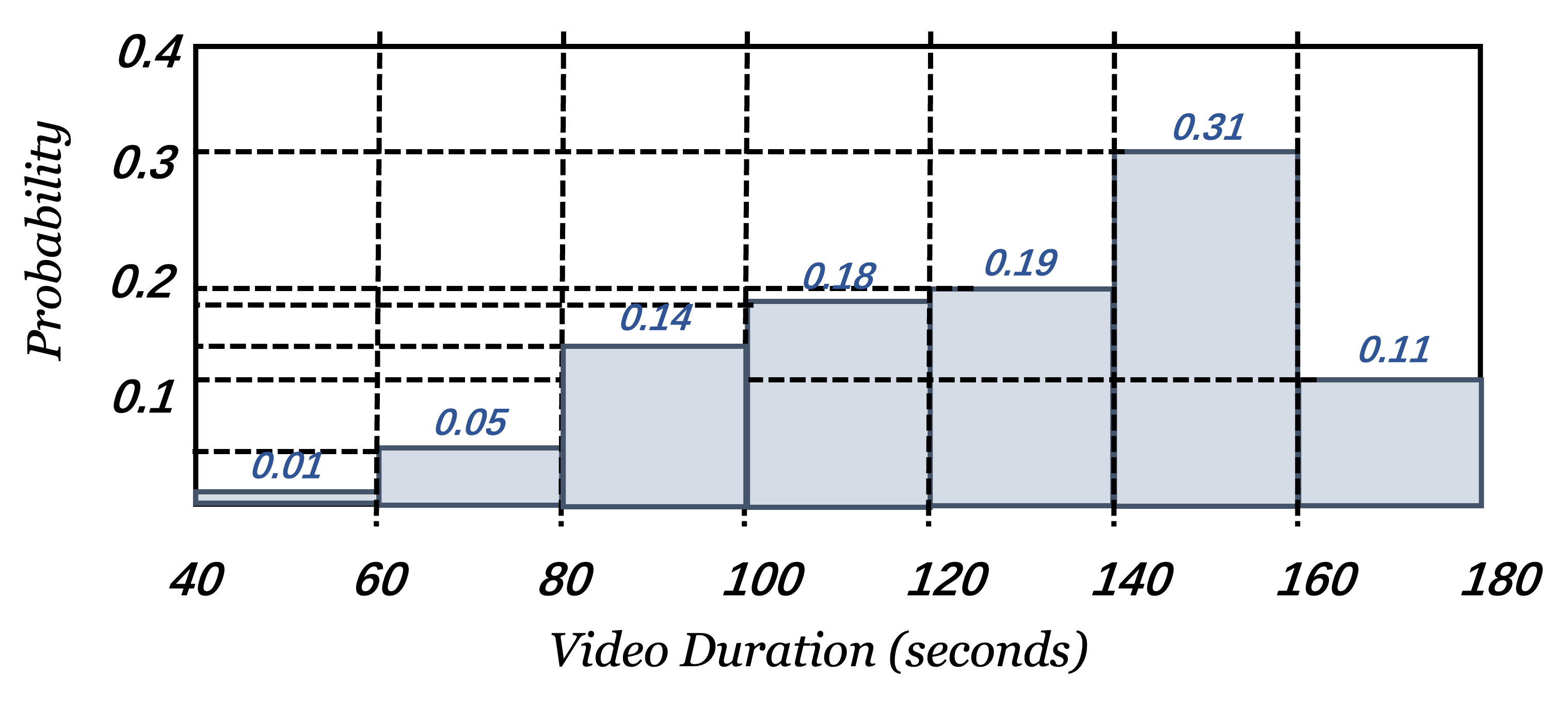}
        \caption{Distribution of video duration.}
        \label{fig:2a}
    \end{subfigure}
    \hfill
    \begin{subfigure}[t]{0.3\textwidth}
        \centering
        \includegraphics[width=\linewidth]{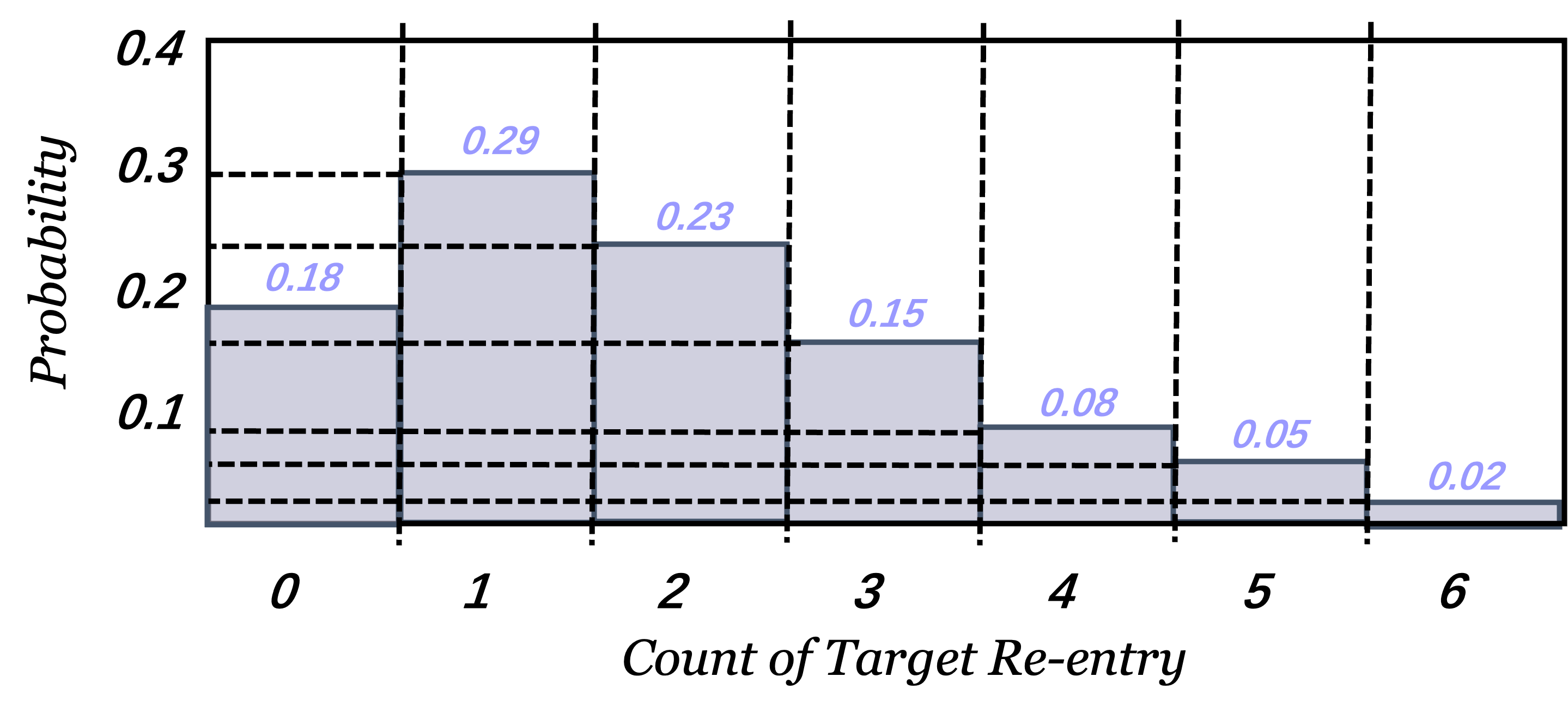}
        \caption{Distribution of re-entry frequency.}
        \label{fig:2b}
    \end{subfigure}
    \hfill
    \begin{subfigure}[t]{0.2\textwidth}
        \centering
        \includegraphics[width=\linewidth]{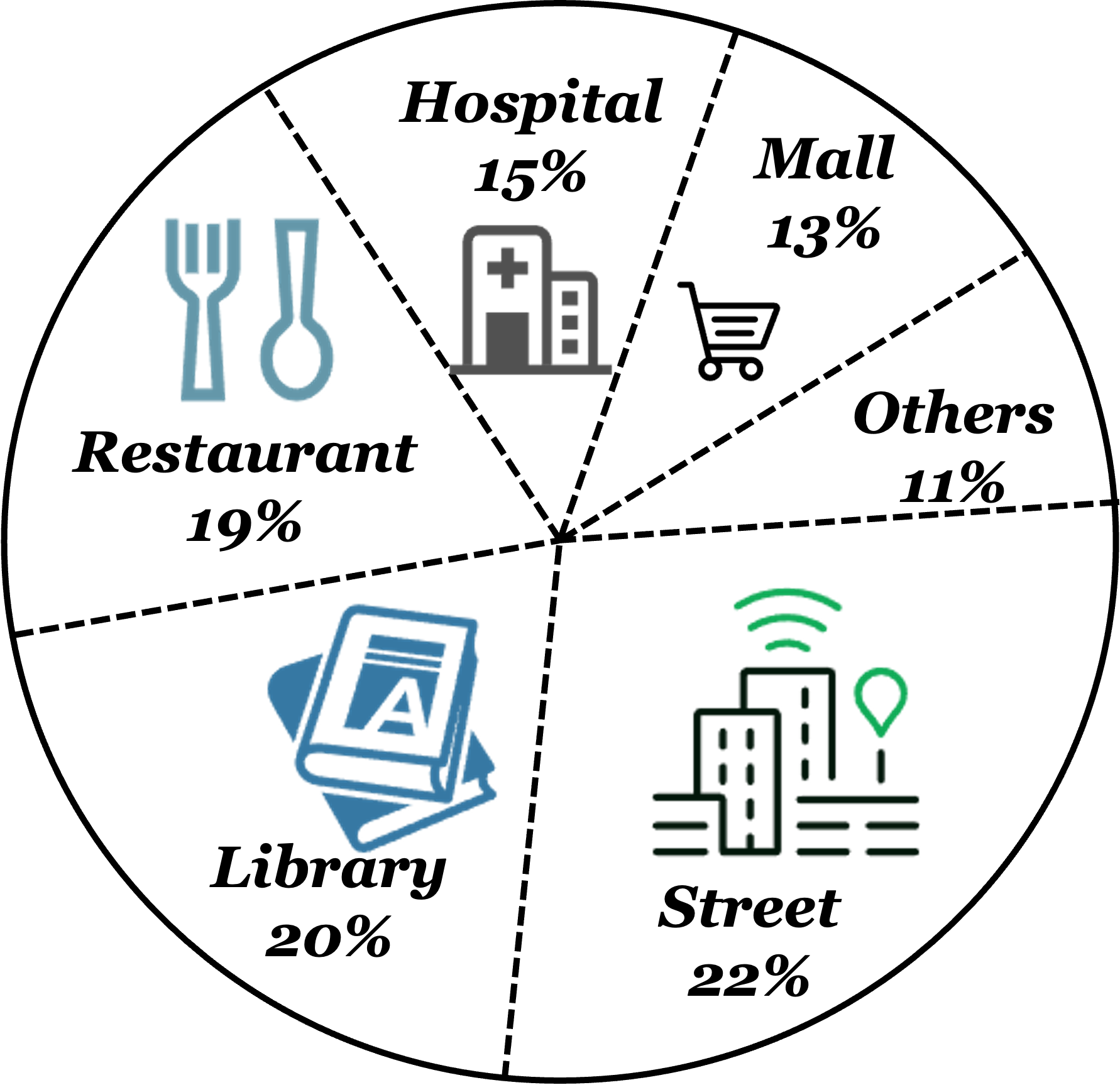}
        \caption{Distribution of scenes.}
        \label{fig:2c}
    \end{subfigure}

    \caption{\textbf{Dataset statistics:} video duration distribution, target re-entry frequency, and scene category composition.}
    \label{fig:fig4}
\end{figure*}
\textbf{Open-Vocabulary Grounding \& Referring.} 
Open-vocabulary detection and referring segmentation enable object localization conditioned on natural language. Grounding DINO series\cite{GroundingDINO,GroundingDINO1.5,DINOv2} tightly integrates language features into the detection head, enabling zero-shot grounding. GLIP\cite{GLIP} improves open-vocabulary generalization via unified language-region pre-training. SAM\cite{SAM}, CLIP\cite{CLIP}, and related models enable promptable mask generation and cross-modal similarity computation.

Referring Video Object Segmentation (R-VOS) aims to localize the language-referred target throughout a video. 
Early works such as URVOS\cite{URVOS} and CMPC\cite{CMPC} mainly perform framewise cross-modal alignment. 
Transformer-based approaches including MTTR\cite{MTTR} and ReferFormer\cite{ReferFormer} enhance long-term semantic reasoning but depend on the referent remaining visible. 
Motion-guided methods like TCE-RVOS\cite{TCE-RVOS} improve short-term propagation under occlusion. 
Recent generative models such as VD-IT\cite{VD-IT} and Ref-Diff\cite{Ref-Diff} improve temporal consistency but lack persistent spatial cues.


In contrast, AR$^2$-4FV anchors the referring expression to persistent background structures in fixed-view scenes, producing an Anchor Map that remains valid during invisibility. This semantic anchor mitigates the long-term inconsistency issues faced by open-vocabulary grounding methods.

\textbf{Long-Term Tracking \& Re-Identification.}
Video tracking aims to maintain target localization and association over time. Most existing works, including FairMOT\cite{FairMOT}, ByteTrack\cite{Bytetrack}, OC-SORT\cite{OC-SORT} and BoT-SORT\cite{BoT-SORT}, focus on short-term settings where the target remains visible. OVTrack~\cite{OVTrack} extends this paradigm to
open-vocabulary tracking by integrating text-based retrieval with association.
Many existing models \cite{GlobalTrack,LCA,Quasi} achieve good performance on long-term benchmarks such as LaSOT \cite{LaSOT} and TAO \cite{TAO}, but they are not designed to handle re-entry scenarios. Therefore, current tracking methods struggle to recover the target after long disappearance. Person ReID models such as OSNet\cite{OSNet}, FastReID\cite{FastReID}, and TransReID\cite{TransReID} provide strong appearance embeddings for identity association, but once the target leaves the scene, appearance-only matching becomes unreliable and tends to be confused by similar distractors.

Consequently, our ReID-Gating strategy leverages structural priors from the Anchor Map and validates identity within the anchor coordinate space, reducing identity drifts near occlusion boundaries and during re-entry.


\textbf{Fixed-View Priors \& Structured Scene Memory.}
Fixed-view videos provide stable background structures that have long been leveraged in background modeling methods such as MOG\cite{MOG}, MOG2\cite{MOG2}, and ViBe\cite{ViBe}. These approaches demonstrate that extracting scene-invariant cues can significantly improve the stability of foreground estimation. Recent advances further show that structured intermediate representations such as spatial maps, segmentation masks, or key points can serve as strong priors for downstream strategy or perception. For example, RoboGround\cite{RoboGround} adopts grounding masks to strategy learning in a two-stage paradigm. 

Inspired by structure-aware perception, AR$^2$-4FV follows a build-then-use paradigm specialized for fixed-view videos. Anchor Map guides re-entry search, restricts proposals to anchor-responsive regions, and supports ReID-Gating directly at the candidate association level which complements existing intermediate-representation approaches.

%% file: sec/3_Dataset.tex
\section{AR$^{2}$-4FV-Bench}
\begin{figure*}
  \centering
  \includegraphics[width=\textwidth]{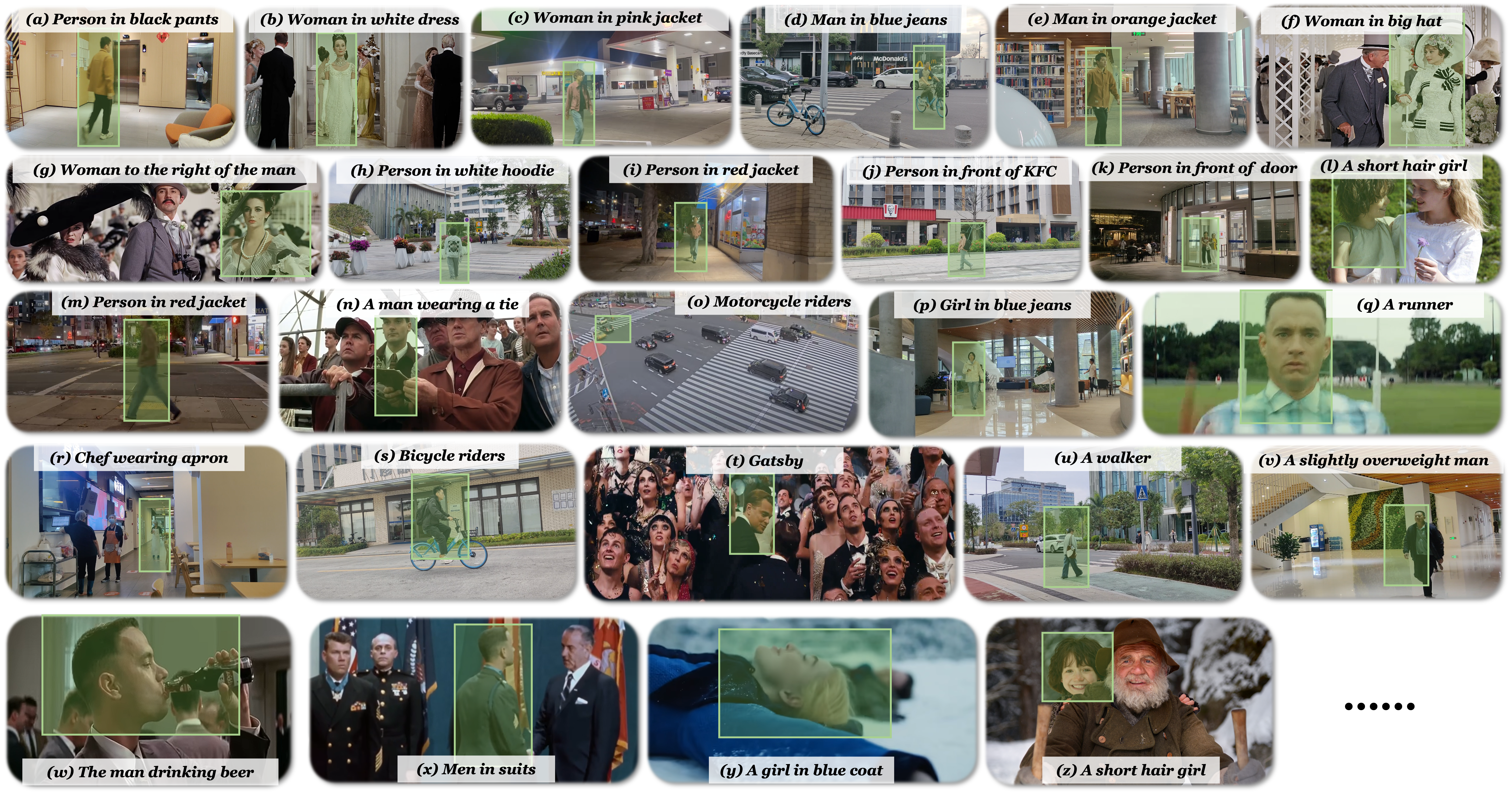}
  \captionof{figure}{\textbf{Diverse language-guided referring queries and annotated referents in AR$^2$-4FV-Bench.} Examples span indoor and outdoor fixed-view scenes as well as cinematic clips. Queries cover anchor-referential descriptions (e.g., “the person near the doorway”) and attribute-based disambiguation by color, clothing, or pose. Green boxes indicate ground-truth referents.}
  \label{fig:fig4}
\end{figure*}
\subsection{Task and Objective}
AR$^{2}$-4FV-Bench focuses on long-term, language-guided referring in fixed-view scenarios.
Given a long video sequence $I_{t=1}^{T}$ captured from a fixed camera viewpoint and a text query $q$, the task is to generate bounding boxes for each frame while maintaining identity consistency throughout occlusion, disappearance, and re-entry.
The dataset setting is closely aligned with the methodology: it does not require the referent to be visible in the first frame and exclude cross-scene outfit changes, while covering intra-scene appearance variations caused by pose, illumination, and scale.
The core objective is to make language-to-scene alignment remain informative even during invisible periods, providing a direct evaluation of the effectiveness of the Anchor Bank, Anchor Map, re-entry prior, and ReID-Gating modules.

\subsection{Collection-Annotation-Query}
\textbf{Collection (Fixed-view, Long-term).}
Cameras are placed on tripods or surveillance positions, keeping the field of view and focal length constant within each video.
We collect 1,684 long videos, each with an average duration exceeding 120 seconds, captured across diverse static scenes such as campus gates, building lobbies, community intersections, and indoor corridors. These videos naturally contain rich real-world dynamics, including daytime to evening illumination shifts, weather-induced lighting variations, moving crowds, shadows, and temporary occlusions.
This setup allows the initial frames to extract scene-invariant elements (supporting the offline Anchor Bank) and produces disappearance and re-entry events over extended periods (supporting online search and re-identification evaluation).

\textbf{Annotation (Visibility, Trajectory, Re-entry).} Framewise visibility is annotated as {\texttt{visible}, \texttt{occluded}, \texttt{absent}}: 
Bounding boxes are provided for \texttt{visible} frames following the evaluation protocol in COCO\cite{COCO};
Each \texttt{absent}→\texttt{visible} transition is explicitly annotated with its timestamp to support TTR and re-capture evaluation.
Some sequences are used to supplement the dataset with distractor trajectories which are similar to the referent. Independent annotations from two annotators are merged, and key boundaries are randomly checked to ensure labeling consistency.

\textbf{Query and Split.}
Each video is accompanied by multiple natural-language queries. The first type is anchor-referential as shown in Figure~\ref{fig:fig4} (e.g., “the man in a gray jacket near the main entrance” or “the cyclist beside the pillar”), which directly triggers language–to-scene alignment during absence periods to form an Anchor Map.
The second type is attribute-based disambiguation (e.g., color, clothing, carried items, etc.) designed to handle multi-instance ambiguity.
Equivalent paraphrases are provided for each referent to evaluate language robustness, while solvable negative examples are introduced during absence periods to prevent false detections on empty frames.
Dataset is partitioned by location to prevent scene leakage and further stratified by the longest disappearance duration and the number of re-entries (Short, Medium, Long; Single or Multiple). Stratified results are reported to separately evaluate the contributions of the re-entry prior (for long absences) and ReID-Gating (for frequent re-entries and similar distractors).

%% file: sec/4_Methodology.tex
\section{Methodology}
\subsection{Overview}
\begin{algorithm}[t]
\small
\caption{AR$^{2}$-4FV}
\label{alg:ar2-4fv}
\SetKwInput{KwReq}{Requires}
\KwIn{frames $\{I_t\}_{t=1}^{T}$ (fixed view), query $q$}
\KwOut{trajectory $\{y_t\}_{t=1}^{T}$}
\KwReq{frozen encoders $f_v,f_l$, alignment heads $\phi_l,\phi_v$, detector $\mathcal{D}$, refiner $\mathcal{R}_{\mathrm{cm}}$, segmentation model SEG, ReID encoder REID, hyper-parameters $\Theta$.}
\BlankLine
\textbf{Offline:}\;
$\mathcal{B} \leftarrow \textsc{BuildBank}(\{I_t\}_{t=1}^{T_0}, f_v)$ \tcp*{Anchor Bank $\{(M_k,p_k,c_k)\}$}
$A \leftarrow \textsc{AnchorMap}(q,\mathcal{B},f_l,\phi_l,\phi_v)$ \tcp*{query-conditioned Anchor Map}
$P^{\mathrm{re}}_0 \leftarrow \textsc{InitPrior}(A)$ \tcp*{$P^{\mathrm{re}}_0(x) \propto A(x)$}
$\mathcal{Q} \leftarrow \varnothing$;\quad $k^\star \leftarrow \mathrm{None}$\;
\BlankLine
\textbf{Online:}\;
\For{$t \leftarrow 1$ \KwTo $T$}{
    $F_t \leftarrow f_v(I_t)$\;
    $\mathcal{R}_t \leftarrow \mathcal{D}(I_t)$\;
    $\tilde{\mathcal{R}}_t \leftarrow \textsc{AnchorGate}(\mathcal{R}_t,A)$ \tcp*{anchor-responsive proposals}
    \If{$|\tilde{\mathcal{R}}_t| = 0$}{
        $P^{\mathrm{re}}_t \leftarrow \textsc{SearchUpdate}(A,P^{\mathrm{re}}_t)$\;
        $y_t \leftarrow \mathrm{None}$;
        
        \textbf{continue}\;
    }
    $B^\star \leftarrow \mathcal{R}_{\mathrm{cm}}(I_t,q,\tilde{\mathcal{R}}_t)$ \tcp*{cross-modal refinement}
    \If{$|B^\star| = 0$}{
        $P^{\mathrm{re}}_t \leftarrow \textsc{SearchUpdate}(A,P^{\mathrm{re}}_t)$\;
        $y_t \leftarrow \mathrm{None}$; 
        
        \textbf{continue}\;
    }
    $r^\dagger \leftarrow \textsc{FusePick}(B^\star,I_t,F_t,A,q,\text{SEG},\phi_l)$ \tcp*{mask-aware fusion}
    \If{$r^\dagger = \varnothing$ \textbf{or} $\textsc{Score}(r^\dagger) < \theta$}{
        $P^{\mathrm{re}}_t \leftarrow \textsc{SearchUpdate}(A,P^{\mathrm{re}}_t)$\;
        $y_t \leftarrow \mathrm{None}$; 
        
        \textbf{continue}\;
    }
    $(\textsf{ok},y_t,h,M_r) \leftarrow \textsc{ReIDGate}(r^\dagger,I_t,A,k^\star,\text{SEG},\text{REID})$ \tcp*{ReID-Gating}
    \If{\textsf{ok} = \textbf{False}}{
        $P^{\mathrm{re}}_t \leftarrow \textsc{SearchUpdate}(A,P^{\mathrm{re}}_t)$\;
        $y_t \leftarrow \mathrm{None}$; 
        
        \textbf{continue}\;
    }
    $\mathcal{Q} \leftarrow \textsc{UpdateQueue}(\mathcal{Q},h)$ \tcp*{momentum identity queue}
    $k^\star \leftarrow \textsc{PickAnchorByIoU}(M_r,\mathcal{B})$ \tcp*{update anchor index}
    $P^{\mathrm{re}}_{t+1} \leftarrow \textsc{RedirectPriorToAnchor}(A,c_{k^\star})$ \tcp*{anchor-based re-entry prior}
}
\Return{$\{y_t\}_{t=1}^{T}$}
\end{algorithm}




\begin{figure*}[!t]
  \centering
  \includegraphics[width=\textwidth]{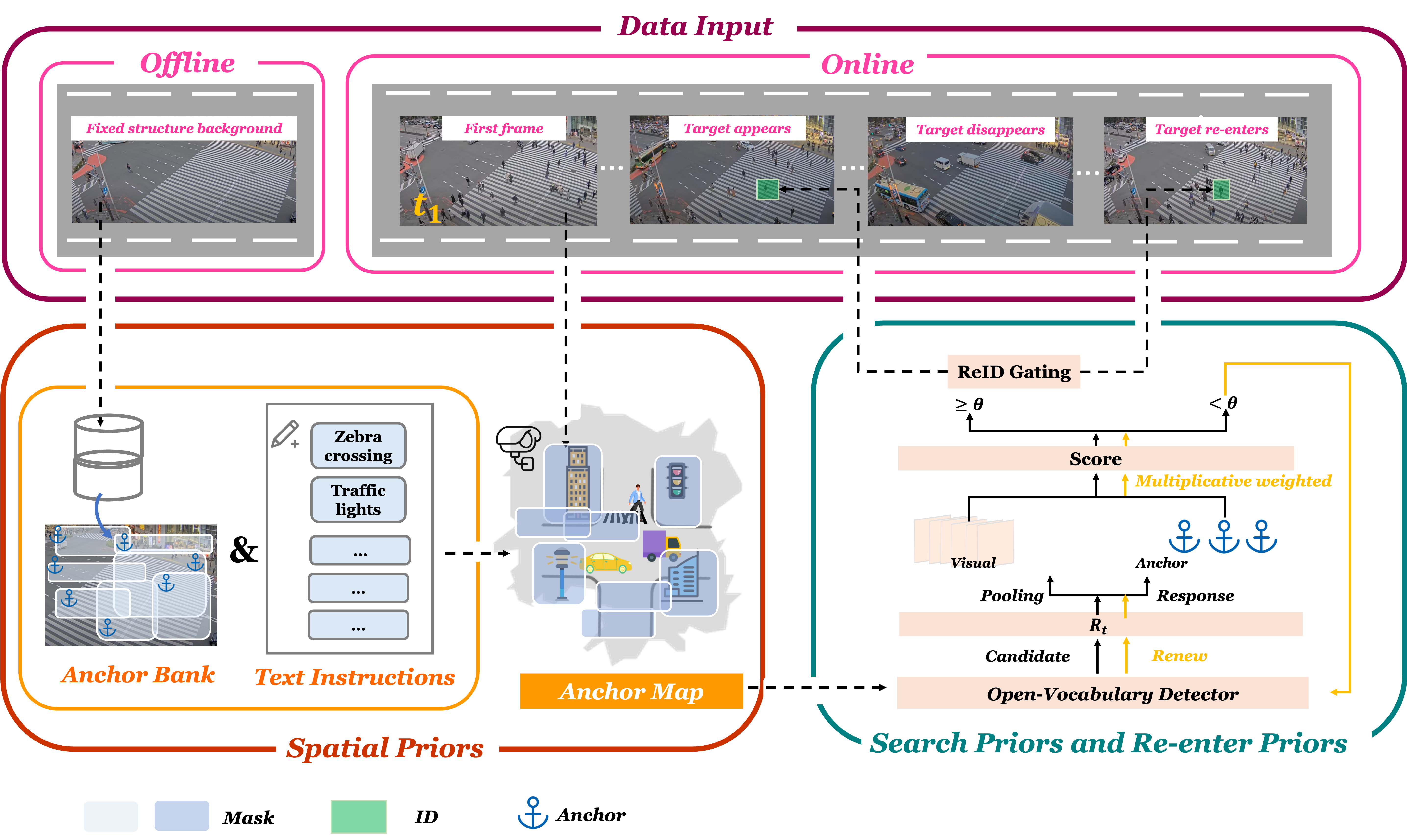}
  \captionof{figure}{\textbf{Overview of AR$^2$-4FV.} Offline: static structures are distilled into an Anchor Bank. Online: the query is aligned to the bank to produce a persistent Anchor Map; this map generates proposals and drives the search and re-entry prior $P^{\mathrm{re}}$. With mask-aware pooling we compute a fusion score, and ReID-Gating validates candidates using appearance similarity, anchor evidence, and displacement in the anchor frame, yielding per-frame boxes $\{y_t\}$.}
  \label{fig:fig3}
\end{figure*}
AR$^{2}$-4FV solves long-term, language-guided referring in fixed-view videos. Given frame sequence $\mathcal{V}=\{I_t\}_{t=1}^{T}$ and a natural-language query $q$, the goal is to output a per-frame trajectory $\{y_t\}$ while preserving identity consistency under occlusion, disappearance, and re-entry. 
First-frame visibility is not assumed and drastic appearance changes are not modeled. 
The pipeline as shown in Figure~\ref{fig:fig3}  and Algorithm~\ref{alg:ar2-4fv} consists of two components: 
(a) \textbf{the language-anchored scene memory }that aligns $q$ to persistent background structure and generates a query-conditioned Anchor Map that remains informative when the referent is invisible (Section~\ref{4.2}); (b) \textbf{the anchor-conditioned association} that restricts proposal sampling to anchor-responsive regions, maintains an anchor-centric re-entry prior, and validates identity with a lightweight ReID-Gating (Section~\ref{4.3}). 
Dense visual features $F_t = f_v(I_t) \in \mathbb{R}^{H\times W\times d_v}$ and a text embedding $e_q = f_l(q) \in\mathbb{R}^{d_l}$ are extracted by frozen encoders.

\subsection{Language-Anchored Scene Memory: Anchor Bank to Anchor Map}
\label{4.2}
\textbf{Anchor Bank (offline).} Fixed cameras induce a stable layout. From the first $T_0$ frames, we synthesize an Anchor Bank by distilling a compact set of anchors $\mathcal{B}=\{(M_k,p_k,c_k)\}_{k=1}^{K}$,
where $M_k\in\{0,1\}^{H\times W}$ is a persistent region mask, $p_k\in\mathbb{R}^{d}$ is prototype of anchor, and $c_k$ is the centroid of anchor. They are pooled on a median-brightness frame $t^\star$:
\begin{equation}
    p_k=\mathrm{Norm}\!\Big(\tfrac{1}{|M_k|}\sum_x M_k(x)\,F_{t^\star}(x)\Big) \in \mathbb{R}^d,
\end{equation}
\begin{equation}
    c_k = \frac{1}{|M_k|}\sum_x M_k(x)x.
\end{equation}
The Anchor Bank is static and provides a scene-aligned coordinate frame.

\textbf{Language-anchor alignment (online).} 
Using lightweight alignment head $\phi_l: \mathbb{R}^{d_l} \rightarrow \mathbb{R}^d, \phi_v: \mathbb{R}^{d_v} \rightarrow \mathbb{R}^d$, the query text $q$ and anchor prototype $p_k$ are mapped into the same subspace:
\begin{equation}
    s_k = \cos\!\big(\phi_l(e_q),\, \phi_v(p_k)\big),
\end{equation}
and obtain the normalized weights by Softmax for each anchor:
\begin{equation}
    \omega_k = \frac{\exp(\tau s_k)}{\sum_{j=1}^{K} \exp(\tau s_j)}.
\end{equation}
By weighting each static region mask, we obtain the Anchor Map:
\begin{equation}
    A(x) = \sum_{k=1}^{K} \omega_k\, M_k(x) \in [0,1].
\end{equation}
Since both $\{M_k\}$ and $\{w_k\}$ remain fixed for a given query, the Anchor Map stays constant during inference and serves as the spatial memory of the query $q$ and the anchor coordinate system.

\subsection{Anchor-Conditioned Association}
\label{4.3}
\textbf{Proposal generation \& cross-modal disambiguation.} An open-vocabulary detector $\mathcal{D}$ proposes regions $\mathcal{R}_t$ on $I_t$. AR$^{2}$-4FV samples only in anchor-responsive areas $\tilde{\mathcal{R}}_t\subseteq\mathcal{R}_t$. We first perform spatial filtering using the response of the Anchor Map: 
\begin{equation}
    \bar A_{bb}(r) = \frac{1}{|\Omega_r^{bb}|} \sum_{x \in \Omega_r^{bb}} A(x),
\end{equation}
\begin{equation}
    \tilde{\mathcal{R}}_t = \{\,\bar A_{bb}(r) \ge \eta\ | r \in \mathcal{R}_t \}\ .
\end{equation}
where \( \Omega_r^{bb} \) denotes the set of pixels within proposal \( r \). The parameter \( \eta \in [0,1] \) is a threshold. A cross-modal refiner $\mathcal{R}_{cm}(I_t,q,\tilde{\mathcal{R}_t})$ to resolve multi-instance ambiguity, generating a filtered bounding box set $B^\star$. 

\textbf{Mask-aware pooling \& fusion scoring.}
For each proposal region \( r \in B^\star \), we generate its binary mask \( M_r \). We denote valid pixels as $\Omega_r^m$ and crop the image within a tight neighborhood region (an expanded bounding box with \(\delta\) pixels):
\begin{equation}
    I^{\text{crop}}_r = (I_t \odot M_r)\big|_{{\text{bbox}(\Omega_r^m)^{+\delta}}}, 
\quad \delta \in \mathbb{N}.
\end{equation}
We derive the mask-aware visual feature $g_v(r)$ by average pooling within refined proposal \(r\):
\begin{equation}
        g_v(r)=\mathrm{Norm}\!\left(\frac{1}{|\Omega_r^m|}\sum_{x \in \Omega_r^m} F_t(x)\right),
\end{equation}
and further compute its average response on the Anchor Map denote as 
\begin{equation}
    \bar{A}_m(r)=
\frac{1}{|\Omega_r^m|}
\sum_{x \in \Omega_r^m} A(x).
\end{equation}

The fusion score couples text–image similarity with anchor evidence:
\begin{equation}
    \mathrm{Score}(r)=\lambda \cos\!\big(g_v(r), g_l(q)\big)+(1-\lambda)\,\bar{A}_m(r),
\end{equation}
where $g_l(q)=\mathrm{Norm}\!\big(\phi_l(e_q)\big)$, $\lambda \in [0,1]$. If $\max_r \mathrm{Score}(r) < \theta$ with $\theta \in [0,1]$, the system enters search mode; otherwise the top candidate enters ReID-Gating stage to be validated. 

\textbf{Search mode \& re-entry prior.} When no reliable candidate is available, we maintain a re-entry prior 
${P}^{\mathrm{re}}_{t} \in [0,1]^{H \times W}$ 
defined in the anchor coordinate frame. 
We initialize it as 
${P}^{\mathrm{re}}_{0} \propto A$, 
and update it using EMA with Gaussian smoothing, followed by $\ell_{1}$ normalization:
\begin{equation}
    \tilde{P}^{\mathrm{re}}_{t}
= \beta \big( G_{\sigma} * \tilde{P}^{\mathrm{re}}_{t-1} \big)
+ (1-\beta) A,
\end{equation}
\begin{equation}
P^{\mathrm{re}}_{t}
= \frac{\tilde{P}^{\mathrm{re}}_{t}}
{\sum_{x} \tilde{P}^{\mathrm{re}}_{t}(x)},
\end{equation}
where $\beta \in [0,1]$ controls temporal smoothing of the re-entry prior.
For the next frame, each candidate receives a multiplicative re-weighting:
\begin{equation}
    W(r) = 
\frac{1}{|M_r|}
\sum_{x} M_r(x)\, A(x)\, P^{\mathrm{re}}_{t}(x),
\end{equation}
\begin{equation}
    \mathrm{Score}(r) \leftarrow \mathrm{Score}(r) \cdot W(r).
\end{equation}

Once the target is confirmed at anchor $k^\star$ (centered at $c^{k^\star}$) at ReID-Gating Stage, 
the prior is redirected toward this anchor:
\begin{equation}
    \tilde{P}^{\mathrm{re}}_{t+1}
= \rho\, G_{\sigma}(\cdot - c^{k^\star})
+ (1-\rho) A,
\end{equation}
\begin{equation}
P^{\mathrm{re}}_{t+1}
= \frac{\tilde{P}^{\mathrm{re}}_{t+1}}
{\sum_{x} \tilde{P}^{\mathrm{re}}_{t+1}(x)},
\end{equation}
where $\rho \in [0,1]$ sets the strength of anchor-based re-direction.

\textbf{ReID-Gating (Identity continuity)}. An image encoder produces an identity embedding $h_v(r)$ on $I^{\mathrm{crop}}_{r}$ and maintains a momentum queue $\mathcal{Q}$, which stabilizes the identity embedding by smoothing frame-to-frame appearance variations. For the to-be-validated candidate,
\begin{equation}
    \mathrm{sim}{\mathrm{ReID}}(r)=\max_{\textbf{q}_j\in\mathcal{Q}}\cos\!\big(h_v(r),\,\textbf{q}_j\big).
\end{equation}
The displacement term is normalized by the image diagonal length 
\begin{equation}
    \hat{\Delta}(r)=\tfrac{\|\mathrm{center}(M_r)-c_{k^\star}\|2}{\sqrt{H^2+W^2}},
\end{equation}
and the gate score is
\begin{equation}
    G(r)=\sigma\!\big(\alpha_1\,\mathrm{sim}{\mathrm{ReID}}(r)\;+\;\alpha_2\,\overline{A}_m(r)\;-\;\alpha_3\,\hat{\Delta}(r)\;+\;b\big),
\end{equation}
where $\alpha_1$, $\alpha_2$, $\alpha_3$ are learned weights for ReID similarity, anchor consistency, and displacement penalty, and $b$ is a bias term.
If \(G(r) \ge \gamma\), the candidate is accepted, and the embedding queue and target index are updated:
\begin{equation}
   \textbf{q} \leftarrow 
\mathrm{Norm}\!\left(
\mu\, \textbf{q} + (1-\mu)\, h_v(r)
\right), 
\end{equation}
\begin{equation}
    k^\star = 
\arg\max_k \mathrm{IoU}\!\left(M_r,\, M_k\right),
\end{equation}
where $\mu \in [0,1]$ is a momentum coefficient that controls the update rate of the identity queue.
\subsection{Implementation}
AR$^{2}$-4FV runs zero-shot with frozen encoders. In our reference instantiation: proposals use an open-vocabulary detector (e.g., GroundingDINO\cite{GroundingDINO}), cross-modal disambiguation uses a RexSeek\cite{RexSeek}-style refiner, masks are produced by a promptable segmentation model (SAM\cite{SAM}), and identity embeddings come from a CLIP\cite{CLIP}-family encoder; a simple NLP pre-parser (spaCy\cite{spacy}) normalizes $q$. Defaults: $K\!=\!64, T_0\!\in\![30,120], \tau\!=\!10, \lambda\!=\!0.6, \theta\!=\!0.4, \beta\!=\!0.8, \gamma\!=\!0.5$. Full training and ablation details are provided in the supplementary materials.

%% file: sec/5_Experiments.tex
\section{Experiments}
\begin{table}[t]
\centering
\caption{Quantitative comparison of AR$^2$-4FV and state-of-the-art models on the AR$^{2}$-4FV-Bench.}
\label{tab:table1}
\begin{tabular}{ccccc}
\toprule[1.5pt]
\toprule
\textbf{Model} & \textbf{Reference} & \textbf{IDF1}$\uparrow$ & \textbf{RCR}$\uparrow$ & \textbf{RCL}$\downarrow$ \\
\midrule
MTTR         & [CVPR'22] & 56.3 & 0.60 & 33.8 \\
ReferFormer & [CVPR'22] & 57.9 & 0.63 & 31.2 \\
OnlineRefer  & [ICCV'23] & 58.6 & 0.64 & 29.9\\
JointNLT  & [CVPR'23] & 59.4 & 0.63 & 30.8 \\
SOC          & [NIPS'23] & 58.7 & 0.64 & 30.3 \\
VideoLISA & [NIPS'24] & 57.2 & 0.60 & 29.5 \\
DsHmp        & [CVPR'24] & 60.4 & 0.66 & 28.6 \\
UVLTrack & [AAAI'24] & 60.8 & 0.66 & 28.2 \\
SSA          & [CVPR'25] & 61.5 & 0.68 & 26.5 \\
DUTrack & [CVPR'25] & 62.3 & 0.69 & 25.8 \\
\midrule
\rowcolor{blue!15}
\textbf{AR$^2$-4FV} & \textbf{-} & \textbf{64.8} & \textbf{0.75} & \textbf{20.1} \\
\bottomrule
\toprule[1.5pt]
\end{tabular}
\end{table}

\begin{table*}[t]
\centering
\caption{Comparison of Precision between AR$^2$-4FV and state-of-the-art models on the AR$^2$-4FV-Bench.}
\label{tab:table2}
\resizebox{0.8\textwidth}{!}{
\begin{tabular}{c c c c c c c c c c c c}
\toprule[1.5pt]
\toprule
\multirow{2}{*}{\textbf{Model}} & \multirow{2}{*}{\textbf{Backbone}} &
\multicolumn{5}{c}{\textbf{Precision}$\uparrow$} & \multirow{2}{*}{\textbf{mAP}$\uparrow$} & \multirow{2}{*}{\textbf{mIoU}$\uparrow$} \\
\cmidrule(lr){3-7}
& &P@0.5 & P@0.6 & P@0.7 & P@0.8 & P@0.9 \\
\midrule
MTTR    &   Video-Swin-T   & 72.1 & 68.0 & 60.3 & 45.6 & 1.2 & 44.6 & 64.1 & \\
ReferFormer & Video-Swin-T &70.5 & 66.9 & 58.8 & 42.7 & 0.9 & 42.8 & 62.9 & \\
OnlineRefer& Video-Swin-B  & 75.4 & 71.4 & 63.6 & 48.5 & 0.8 & 46.1 & 64.2 & \\
JointNLT  & ResNet-50  & 74.2 & 69.8 & 59.5 & 42.1 &0.8 &43.5&61.9& \\
SOC   &  Video-Swin-T     & 69.2 & 65.8 & 57.2 & 41.5 & 0.7 & 41.7 & 61.6 & \\
VideoLISA  & ViT-H & 74.8 & 70.5 & 61.6 & 44.9 & 1.3 & 45.5 &62.1 & \\
DsHmp &   Video-Swin-T    & 72.6 & 69.1 & 60.4 & 44.3 & 1.4 & 45.0 & 63.8 & \\
UVLTrack &  ViT-B  & 75.8 & 70.5 & 61.3 & 45.1 & 1.5&45.8&63.5& \\
SSA  &  CLIP     & 73.6 & 70.3 & 62.4 & 47.1 & 1.4 & 45.2 & 64.0 & \\
DUTrack & ViT-B & 76.5 & 71.2 & 62.8 & 44.5 & 1.5 &46.5&63.7& \\
\midrule
\rowcolor{blue!15}
\textbf{AR$^2$-4FV} &  \textbf{CLIP} & \textbf{77.3} & \textbf{72.2} & \textbf{64.1} & \textbf{49.2} & \textbf{2.6} & \textbf{49.2} & \textbf{66.9} & \\
\bottomrule
\toprule[1.5pt]
\end{tabular}
}
\end{table*}

\subsection{Metrics}
We evaluate fixed-view long-term referring and ReID using a multi-dimensional metric suite that captures spatial localization accuracy, temporal identity continuity, and re-capture robustness. Spatial performance is assessed using mIoU, the standard metric for bounding-box localization. Meanwhile, mAP integrates Precision and Recall across different IoU thresholds $\tau$ to assess detection capability. The threshold $\tau$ represents the overlap ratio, typically ranging from 0.5 to 0.95 with an interval of 0.05 following the evaluation protocol in COCO\cite{COCO}, and the final results are averaged over all thresholds. For temporal consistency in long-term scenarios, we employ IDF1 to evaluate the model’s ability to maintain consistent identity assignment throughout the video sequence. 

Furthermore, to quantify the model’s robustness in handling multiple disappearance and re-entry events, we introduce two specific metrics: Re-capture Rate (RCR) and Re-capture Latency (RCL). RCR measures the proportion of correctly identified targets after re-entering the scene, defined as:
\begin{equation}
    RCR = \frac{1}{K} \sum_{k=1}^{K} \mathds{1} [IoU_k \ge \tau],
\end{equation}
where $K$ denotes the total number of re-entry events in the video. 
RCL evaluates the average frame latency required for the model to successfully re-capture the target after re-entry, defined as:
\begin{equation}
   RCL = \frac{1}{K} \sum_{k=1}^{K} (t_{det,k} - t_{re,k}), 
\end{equation}
where $t_{re,k}$ is the frame index of the $k$-th re-entry and $t_{det,k}$ is the first frame where the target is correctly detected after re-entry. 
Together, these metrics provide a comprehensive evaluation framework to assess the AR$^2$-4FV model from spatial, temporal, and re-entry perspectives under fixed-view video settings.

\begin{table*}[t]
\centering
\caption{Ablation study of AR$^2$-4FV with component-wise configurations.}
\label{tab:table3}
\resizebox{0.8\textwidth}{!}{%
\begin{tabular}{ccccccccc}
\toprule[1.5pt]
\toprule
\multicolumn{4}{c}{\textbf{Components}} & 
\multirow{2}{*}{\textbf{mIoU}$\uparrow$} & 
\multirow{2}{*}{\textbf{mAP}$\uparrow$} & 
\multirow{2}{*}{\textbf{IDF1}$\uparrow$} & 
\multirow{2}{*}{\textbf{RCR}$\uparrow$} & 
\multirow{2}{*}{\textbf{RCL}$\downarrow$} \\
\cmidrule(lr){1-4}
\textbf{Baseline} & \textbf{Anchor Map} & \textbf{Re-ID Gating} & \textbf{Re-entry Prior} &  &  &  &  &  \\
\midrule
\checkmark &            &            &            & 63.2  &  45.2 &  61.2 & 0.67 & 27.1  \\
\checkmark & \checkmark &            &            & -  &  - &  - & -  &  - \\
\checkmark & \checkmark & \checkmark &      & 64.7  &  46.3 & 62.2  &  0.70 & 26.9 \\
\checkmark & \checkmark &  &   \checkmark   & 63.8  &  45.5 & 61.3  &  0.68 & 21.3 \\
\rowcolor{blue!15}
\checkmark & \checkmark & \checkmark & \checkmark & \textbf{66.9} & \textbf{49.2}  & \textbf{64.8} & \textbf{0.75}  & \textbf{20.1}  \\
\bottomrule
\toprule[1.5pt]
\end{tabular}%
}
\end{table*}
\subsection{Main Results}
To comprehensively evaluate the effectiveness of the proposed method, we compare AR²-4FV with several representative models that cover the main paradigms of language referring in fixed-view video understanding.
All models are trained and tested on the AR$^{2}$-4FV-Bench to ensure fair comparison.
All results are evaluated under same input configurations and inference settings.

\textbf{Models.}
\begin{itemize}
    \item \textit{MTTR\cite{MTTR}}: An end-to-end multimodal Transformer that jointly models video and text for temporally consistent segmentation and tracking.
    \item \textit{ReferFormer\cite{ReferFormer}}: A Transformer that uses language embeddings as queries for unified cross-frame language–vision modeling and consistent target segmentation.
    \item \textit{OnlineRefer\cite{OnlineRefer}}: A language-guided video segmentation model with inter-frame memory enabling real-time, high-accuracy tracking and segmentation.
    \item \textit{SOC\cite{SOC}}: A video-level semantic aggregation R-VOS model that aligns visual and linguistic features via semantic-guided clustering to enhance target understanding and segmentation accuracy.
    \item \textit{DsHmp\cite{DsHmp}}: A language-guided video segmentation model using dual-space hybrid prompting to enhance cross-frame consistency and fine-grained segmentation.
    \item \textit{SSA\cite{SSA}}: A R-VOS framework that enhances visual–language grounding through semantic alignment with CLIP and improves temporal consistency by aligning object trajectories across frames.
\end{itemize}
\textbf{Analysis.} Compared with existing R-VOS models, our method performs best on the AR$^{2}$-4FV-Bench. ReferFormer and MTTR rely on motion cues and fail under minimal appearance changes; The video-level semantic aggregation of SOC is not effective in static scenes; OnlineRefer accumulates errors without stable motion updates; and DsHmp overfits weak temporal cues in fixed views, SSA relys on short-term appearance and query-driven local alignment. Overall, these models suit dynamic scenes but fail to exploit the stable spatial anchors and long-term consistency of fixed-view videos.

In Table~\ref{tab:table1}, AR$^2$-4FV outperforms the best baseline by +10.3\% RCR and -24.2\% RCL. For localization performance in Table~\ref{tab:table2}, AR$^2$-4FV achieves improvements of +6.7\% in mAP and +4.2\% in mIoU over the best baseline. Overall, these improvements prove that coupling language with fixed-view scene structure and using anchor-based re-entry priors substantially strengthens identity continuity, re-capture reliability, and spatial localization accuracy.


\subsection{Ablation Studies}
To assess the contribution of each key component, we conduct a stepwise ablation study on AR$^2$-4FV, examining the Anchor Map, ReID-Gating, and re-entry Prior under identical training and inference settings (Table~\ref{tab:table3}). ReID-Gating further improves IDF1 by mitigating identity drift over long sequences, while the re-entry Prior enhances RCR and reduces RCL, enabling faster and more reliable re-capture when the target returns.

Overall, the Anchor Map ensures stable spatial grounding, the ReID-Gating maintains identity continuity, and the re-entry Prior enhances the efficiency of re-capture. The integration of these modules enables the model to maintain spatio-temporal consistency and achieve robust re-capture performance in fixed-view videos.

%% file: sec/6_Conclusion.tex
\section{Conclusion}
We propose AR$^2$-4FV, a framework for long-term language-guided referring and re-identification in fixed-view videos. 
AR$^2$-4FV leverages stable background structures to construct an Anchor Bank and generate an Anchor Map that serves as a persistent language–scene memory. It further integrates re-entry priors and ReID-Gating to maintain identity continuity during disappearance and re-entry.
Extensive experiments demonstrate that AR$^2$-4FV achieves stronger identity consistency and faster re-capture. We also introduce AR$^2$-4FV-Bench, the first benchmark featuring fixed-view videos with explicit occlusion, absence and re-entry annotations. Together, the framework and benchmark establish a solid foundation for robust long-term referring and re-identification stable environments.

%% file: sec/7_Acknowledgement.tex
\section{Acknowledgement}

This work was supported by Guangdong Provincial Project (Grant No. 2024QN11X053) and by the Youth S\&T Talent Support Programme of GDSTA (Grant No. SKXRC2025468).